\PassOptionsToPackage{table,xcdraw,dvipsnames}{xcolor}

\documentclass[10pt,twocolumn,letterpaper]{article}

\usepackage{cvpr}              

\usepackage{graphicx}
\usepackage{amsmath}
\usepackage{amssymb}
\usepackage{booktabs}

\definecolor{cvprblue}{rgb}{0.21,0.49,0.74}
\usepackage[pagebackref,breaklinks,colorlinks,allcolors=cvprblue]{hyperref}

\usepackage{url}            
\usepackage{amsfonts}       
\usepackage{nicefrac}       
\usepackage{microtype}      

\usepackage{makecell}
\usepackage{amsthm}
\usepackage{bm}
\usepackage[table,xcdraw]{xcolor}
\usepackage{multicol}
\usepackage{colortbl}
\usepackage{multirow}
\usepackage{enumitem}
\usepackage{bbding}
\usepackage{color}

\usepackage{algorithm}
\usepackage{listings}
\usepackage{natbib}
\usepackage[misc]{ifsym}

\usepackage{etoolbox}
\makeatletter
\AfterEndEnvironment{algorithm}{\let\@algcomment\relax}
\AtEndEnvironment{algorithm}{\kern2pt\hrule\relax\vskip3pt\@algcomment}
\let\@algcomment\relax
\newcommand\algcomment[1]{\def\@algcomment{\footnotesize#1}}
\renewcommand\fs@ruled{\def\@fs@cfont{\bfseries}\let\@fs@capt\floatc@ruled
  \def\@fs@pre{\hrule height.8pt depth0pt \kern2pt}%
  \def\@fs@post{}%
  \def\@fs@mid{\kern2pt\hrule\kern2pt}%
  \let\@fs@iftopcapt\iftrue}
\makeatother

\definecolor{ForestGreen}{RGB}{34,139,34}

\renewcommand{\paragraph}[1]{\medskip\noindent\textbf{#1.~}}

\newcommand{\dak}[1]{\left\{#1\right\}}

\newcommand{\argmin}[1]{{\mathop{\arg\mathrm{min}}_{#1}\,}}

\newcommand{\bmb}{\bm{b}}

\newcommand{\bmv}{\bm{v}}

\newcommand{\bmI}{\bm{I}}

\newcommand{\bmV}{\bm{V}}

\newcommand{\bmZ}{\bm{Z}}

\newcommand{\calC}{\mathcal{C}}

\newcommand{\calL}{\mathcal{L}}

\newcommand{\bbR}{\mathbb{R}}

\newcommand{\modelname}{AutoSSVH}

\newcounter{qcounter}

\def\paperID{996} 
\def\confName{CVPR}
\def\confYear{2025}

\title{\modelname{}: Exploring Automated Frame Sampling for \\Efficient Self-Supervised Video Hashing }

\author{
Niu Lian$^{1}\thanks{These authors contributed equally to this work.}$ , \ 
Jun Li$^{1}{}^{*}$, \ 
Jinpeng Wang$^{2*}\thanks{Corresponding authors.}$ , \ 
Ruisheng Luo$^{2}$, \ 
Yaowei Wang$^{3}$, \ 
Shu-Tao Xia$^{2,3}$, \ 
Bin Chen$^{1}{}^{\dag}$ \\
$^1$Harbin Institute of Technology, Shenzhen \\
$^2$Tsinghua Shenzhen International Graduate School, Tsinghua University \\
$^3$Research Center of Artificial Intelligence, Peng Cheng Laboratory \\
{\small \tt\{220110904,220110924\}@stu.hit.edu.cn}\\
{\small \Letter\ \tt \{wjp20@mails.tsinghua.edu.cn, chenbin2021@hit.edu.cn\}} \\
}

\begin{document}
\maketitle
\begin{abstract}
Self-Supervised Video Hashing (SSVH) compresses videos into hash codes for efficient indexing and retrieval using unlabeled training videos.  
Existing approaches rely on random frame sampling to learn video features and treat all frames equally.  
This results in suboptimal hash codes, as it ignores frame-specific information density and reconstruction difficulty.  
To address this limitation, we propose a new framework, termed \textbf{\modelname{}}, that employs adversarial frame sampling with hash-based contrastive learning.  
Our adversarial sampling strategy automatically identifies and selects challenging frames with richer information for reconstruction, enhancing encoding capability.  
Additionally, we introduce a hash component voting strategy and a point-to-set (P2Set) hash-based contrastive objective, which help capture complex inter-video semantic relationships in the Hamming space and improve the discriminability of learned hash codes.  
Extensive experiments demonstrate that \modelname{} achieves superior retrieval efficacy and efficiency compared to state-of-the-art approaches.  
Code is available at \url{https://github.com/EliSpectre/CVPR25-AutoSSVH}.

\end{abstract}    
\section{Introduction}
\label{sec:intro}

\begin{figure}[h]
    \centering
    \begin{subfigure}[t]{0.45\textwidth}
        \includegraphics[width=\textwidth]{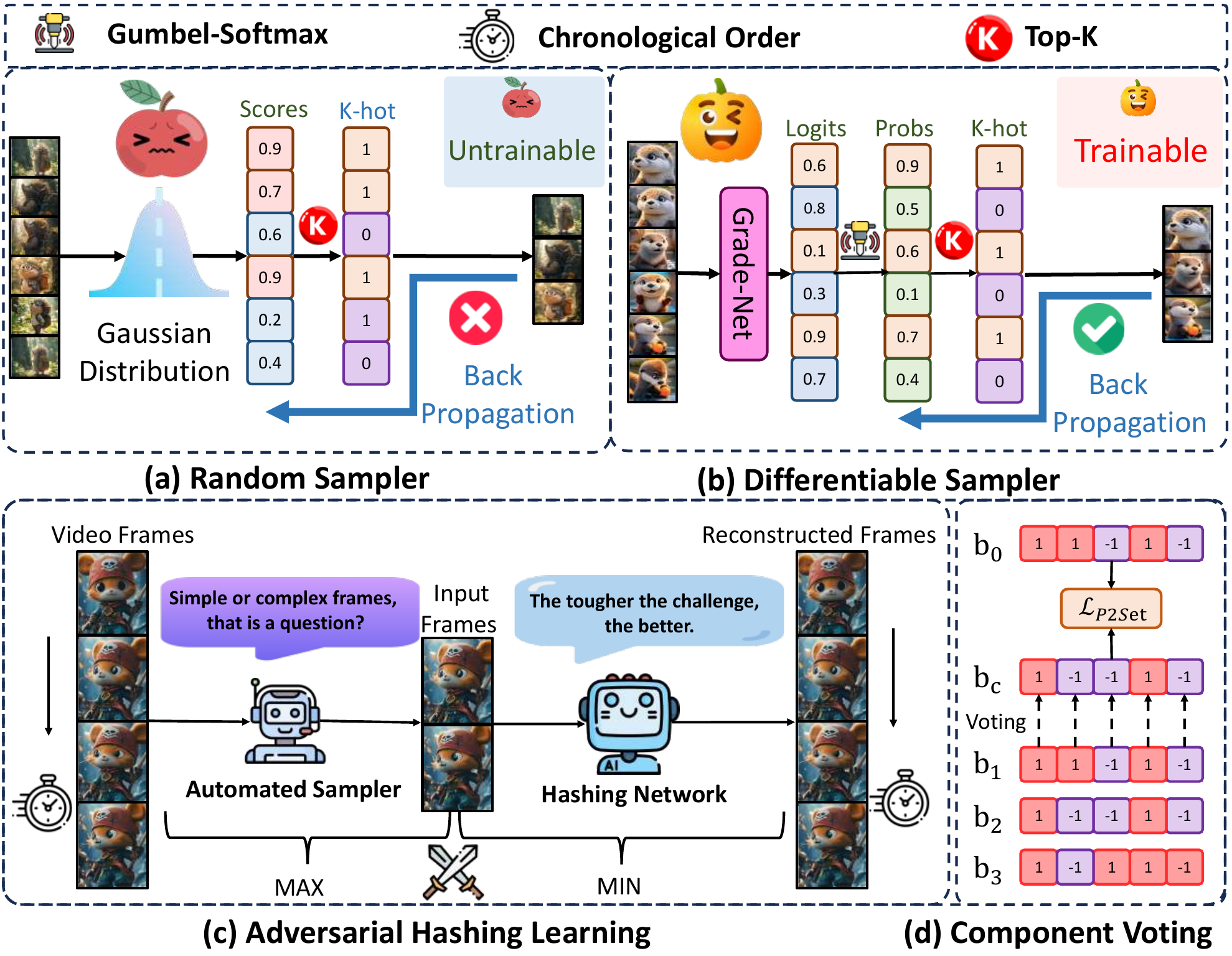}
    \end{subfigure}
    \caption{(a) Existing methods treat all frames equally and randomly sample frames from the video.  (b) In contrast, our approach leverages the Gumbel-Softmax technique to achieve differentiable frame sampling. 
    (c) We propose a GAN-based framework for hash learning, where the frame sampler tries to maximize learning objectives and the hashing network learns to minimize. 
    (d) We further derive hash anchors via a component voting strategy, which supplements global semantic information and enhances hash learning.}
    \label{fig:Intro}
\end{figure}

Content-based video retrieval \cite{CBVH1,CBVH2,CBVH3} is vital in scenarios such as digital forensics and video surveillance systems, where it is used to identify and retrieve specific videos based on visual content, aiding in evidence analysis and investigation. However, because videos usually exist in high-dimensional space, retrieval is not only time-consuming but also demands significant computational resources. Hashing techniques, which map high-dimensional data to low-dimensional representations, greatly reduce both computation and storage requirements, and bitwise operations provide fast processing, making them widely used for image and video retrieval. This paper focuses on unsupervised video hashing retrieval to alleviate the need for manual annotation.

Compared to 2D image retrieval, video hash retrieval presents greater challenges, as it necessitates the modeling of temporal dynamics and the intricate dependencies that exist between videos. Most existing methods design frame reconstruction tasks to enhance the semantic information of hash codes. Early approaches, such as SSVH \cite{SSVH}, BTH \cite{BTH}, involved passing the entire video through the network and reconstructing all the frames. Later, ConMH \cite{ConMH} introduced a mask auto-encoder (MAE \cite{MAE_2022}), which randomly sampled a subset of frames for reconstruction. However, as depicted in \Cref{fig:Intro}(a), random sampler methods \cite{ConMH, S5VH}, overlook the information density of different frames by treating all frames equally, resulting in suboptimal hash codes, thereby degrading the performance of video retrieval.

To address the ongoing issues, we propose \modelname{}, an innovative hash learning framework with \textbf{automated frame sampling}. 
As illustrated in \Cref{fig:Intro}(b), we design a Grade-Net that assigns a score to each frame, utilizing a differentiable Top-K frame sampling mechanism in conjunction with adversarial learning. The overall framework is depicted in \Cref{fig:Intro}(c). Specifically, while the sampler aims to increase the complexity of the reconstruction task, the hashing network simultaneously optimizes its capability to generate hash codes with richer semantic information. The two components form an adversarial hashing learning framework, engaging in a Min-Max adversarial game, where they mutually constrain and evolve in tandem. Ultimately, through Automated Sampling, the model autonomously identifies and processes more challenging frames, optimizing them within its self-imposed difficulties. This dynamic interaction enhances \modelname{}'s capacity to address complex cases and generate higher-quality hash codes.

Adversarial training effectively enables the identification of challenging key frames with more information. However, it is accompanied by a slowdown in the convergence rate of the training process. To address this issue, we introduce a \textbf{point-to-set (P2Set) hash-based contrastive objective},  which accelerates the convergence of adversarial training and captures global high-level semantics. More precisely, first, outlined in \Cref{fig:Intro}(d), we apply Component Voting to obtain an anchor code for each semantic cluster, and then leverage P2Set hash contrastive learning to minimize the distance between the hash code of each video view and its corresponding anchor code. The comparison results presented in \Cref{fig:comparative analysis} substantiate the efficacy of P2Set hash-based contrastive learning in optimizing the training dynamics.

The primary contributions can be summarized as follows: 
\begin{itemize}
    \item[$\bullet$] We propose an adversarial strategy-based automated sampling method for mining hard frames in videos, which captures frame reconstruction difficulty and selects challenging frames to enhance the model's encoding capability.
    \item[$\bullet$] We introduce a P2Set hash contrastive learning that incorporates component voting, which facilitates global-level information aggregation, allowing the hash code to effectively encode comprehensive neighborhood relationships.
    \item[$\bullet$] We conduct extensive experiments on four benchmark datasets: ActivityNet \cite{Activitynet}, FCVID \cite{FCVID}, UCF101 \cite{UCF} and HMDB51 \cite{HMDB}, demonstrating the effectiveness and high efficiency of our proposed approach.
\end{itemize}

\section{Related Works}
\label{sec:related_works}

\subsection{Self-Supervised Video Hashing}\label{subsec:related_ssvh}

 Video hashing methods are designed to compress videos into binary hash codes, thereby  enhancing both the efficiency and accuracy of video retrieval systems. 
  Previous self-supervised video hashing methods like  MPH \cite{MPH} and spectral hashing \cite{spectral_hashing}, relied on image hashing techniques, treating a video as a collection of independent frames. These approaches overlooked the temporal dependencies inherent in video data, leading to suboptimal retrieval performance. 

To overcome the complexity of temporal information and the lack of labeled data ,  a series of enhanced methods have been proposed. VHDT \cite{VHDT}  was the first approach to incorporate the video structure.
To reduce training costs,  MCMSH \cite{MCMSH}, which based on a lightweight MLP-Mixer \cite{MLPMixer_2021} architecture, captured temporal information through long, medium, and short-range distances. Inspired by Bert \cite{BERT}, BTH \cite{BTH}  was proposed for bidirectional temporal information capture. Additionally,  due to the high-dimensional nature of videos, SSTH \cite{SSTH} and SSVH \cite{SSVH} used  K-means clustering to generate pseudo-labels, capturing neighborhood information. ConMH \cite{ConMH}  applied MAE \cite{MAE_2022} and  contrastive learning \cite{he2020momentum} to achieve good performance. CHAIN \cite{CHAIN}constructed  Frame Order Verification and prototypical contrastive learning to adjust the model's perception of videos, while BerVAE \cite{BerVAE}  employed an enhanced Bernoulli Variational Auto-Encoder to generate corresponding hash codes.

Although approaches vary, they typically use a unified frame sampling algorithm, most based on Gaussian random sampling. However, due to varying information content and reconstruction difficulty across frames, random sampling fails to identify and prioritize the key frames essential for effective reconstruction, leading to suboptimal hash codes. To address this, we propose an adversarial strategy for automatic hard-frame mining, focusing on frames with higher reconstruction difficulty to improve feature extraction. Additionally, we introduce a component-voting-based component voting strategy to capture higher-level semantics, enhancing retrieval performance and accelerating the convergence of the training process.

\subsection{\textbf{Sampling Strategy in Vision Transformer}}

The sampling strategy in Vision Transformers \cite{vit} is primarily manifested in the Masked Image Modeling (MIM) task, which involves various masking strategies. Early MIM approaches, such as MAE and Video MAE \cite{videomae}, typically relied on random sampling to select patches. However, this random sampling approach often reduced the challenge of self-supervised learning, resulting in suboptimal performance. In response, researchers have proposed a variety of more sophisticated sampling strategies. For instance, AttMask \cite{attmask} introduced an attention-guided sampling method, where the selection of patches is directed by the attention map. HPM \cite{hpm} adopted a teacher-student framework, where the teacher model predicts the reconstruction loss for each patch, thus guiding the student model's sampling process. SemMAE \cite{semmae} implemented a semantic-based masking strategy by leveraging semantic information learned through the Vision Transformer. AdaMAE \cite{bandara2023adamae} employed a policy gradient algorithm from reinforcement learning to guide token sampling. ADIOS \cite{adios} combined MIM with adversarial training, jointly training both the generator and discriminator through adversarial learning.
However, most sampling methods either rely on another powerful model or require multi-stage adversarial training, as in the case of GANs \cite{gan}.  The adversarial automated sampler proposed in this work employs a lightweight Grade-Net for frame scoring. It utilizes the Gumbel-Softmax operation \cite{gumbel} to enable differentiable Top-K selection, thereby facilitating gradient propagation. Adversarial training is conducted in a single stage through gradient reversal \cite{grl},  improving both the operational efficiency and temporal speed of the video sampling process.

\section{The Proposed Approach}
\label{sec:method}
\begin{figure*}[t]
    \centering
    \includegraphics[width=\textwidth]{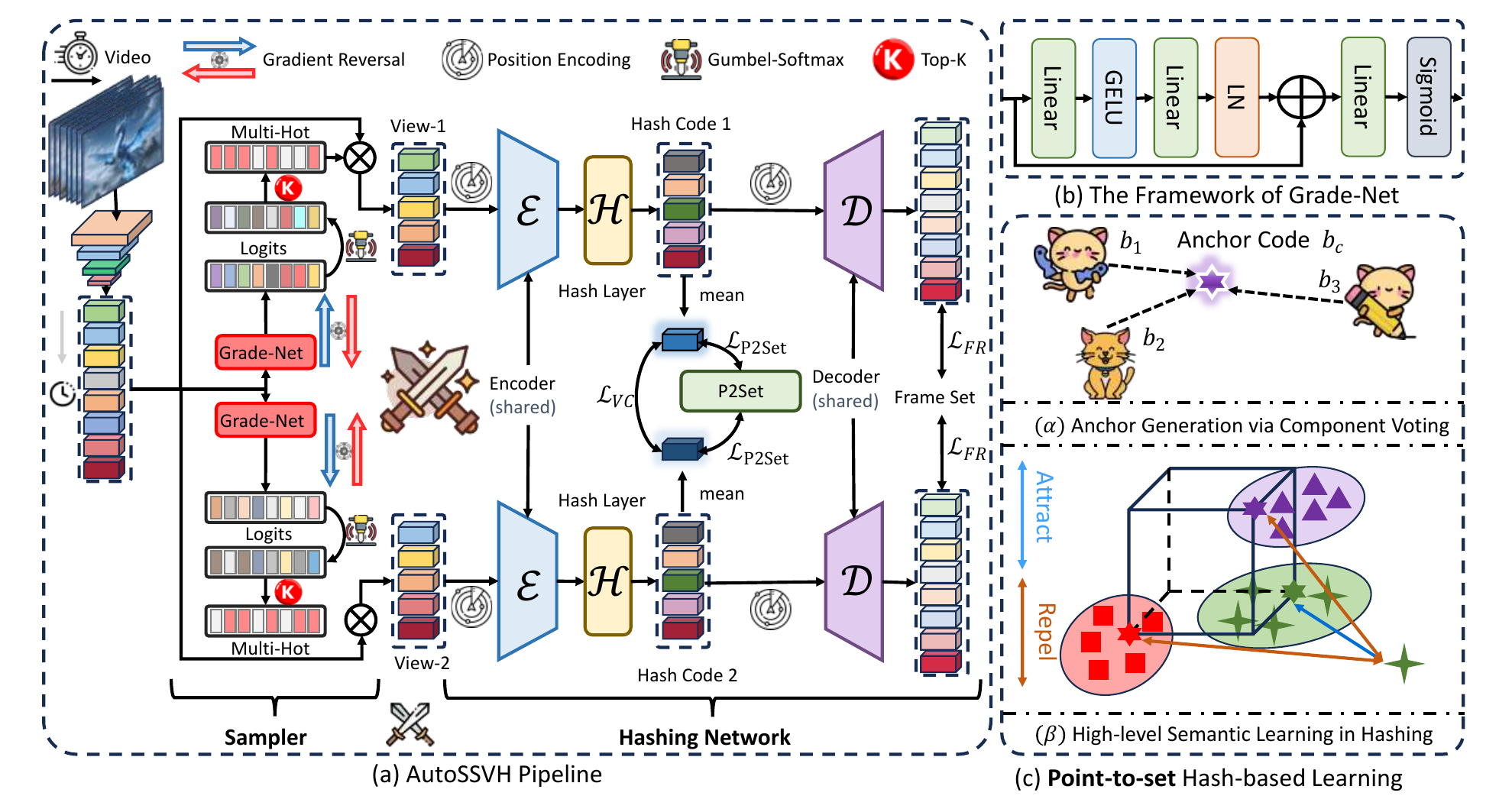}
    \caption{Our Proposed \modelname{}.
    (\textbf{a}) \modelname{} Pipeline. \modelname{} leverages an adversarially-guided automated sampler, which utilizes the Gumbel-Softmax TopK operation and gradient reversal to select frames exhibiting high reconstruction difficulty within a video. This process generates two sequences with reduced informational content, which are subsequently processed by the hashing network to generate hash codes $\bm{b_i}$ and $\bm{b_j}$ for view contrast learning. $\mathcal{L}_\mathsf{VC}$ is then computed based on these hash codes and those from other sequences. The encoder generates hash codes for the entire training set, followed by pseudo-labeling via k-means clustering. Component voting is then applied to determine cluster centers. Point-to-set (P2Set) hash-based learning is performed next, with  $\mathcal{L}_\mathsf{P2Set}$ computed accordingly. Finally, the video is reconstructed, and the frame reconstruction loss $\mathcal{L}_\mathsf{FR}$ is evaluated.
    (\textbf{b})  The Framework of Grade-Net. (\textbf{c})   Point-to-set Hash-based Learning. ($\alpha$) Anchor Generation via Component Voting ($\beta$) High-level Semantic Learning in Hashing.
}
    \label{fig:arc}
\end{figure*}

\subsection{Preliminaries and Overview} \label{subsec:Preliminaroes}

Consider an unlabeled video dataset $\calC=\dak{\bmV_i}_{i=1}^{N}$, where $\bmV_i\in\bbR^{M_0\times D}$ denotes frame features of the $i$-th video, extracted by pre-trained 2D CNNs \cite{vgg,ResNet}. 
Here $M_0$ is the number of frames in each video, and $D$ denotes the dimensionality of the feature vector for each frame.
The goal of self-supervised video hashing (SSVH) is to map $\bm{V}_i$ to a $K$-bit hash code vector $\bm{b}_i \in \{-1, +1\}^K$ in the Hamming space.
In this paper, we propose \textbf{\modelname{}}, a Transformer-based hashing network trained with an automated frame mask sampler, as illustrated in \Cref{fig:arc}.

\subsection{Differentiable Frame Mask Sampler} \label{subsec:sampler} 

\paragraph{Grade-Net}
As illustrated in \Cref{fig:arc}(b), Grade-Net is composed of a lightweight MLP layer, which is employed to assign scores to each input frame. 
Given all frame features of the $i$-th video $\bm{V}_i \in \mathbb{R}^{M_0 \times D}$, we can obtain the score for each frame as follows:
\begin{equation}
\bm{S}_i=f\left(W_4 \left(\text{LN} \left(W_2 \left( \sigma\left(W_1\bm{V}_i\right)\right)\right)+ \bm{V}_i \right)\right) \in \mathbb{R}^{M_0 \times 1},
\end{equation}
where $W$ denotes linear layer, $\sigma(\cdot)$ represents  GELU function, $\text{LN}(\cdot)$ is LayerNorm, $f(\cdot)$ equals sigmoid function.

\paragraph{Gumbel-Softmax TopK Sampling}
The Gumbel-Softmax operation facilitates differentiable selection of target frames from discrete samples. Specifically, we employ a straight-through estimator (STE). During the forward pass, a multi-hot vector is generated using the TopK operation based on the probability distribution to select $k$ frames. 
In the backward pass, gradients are computed using the output of the Gumbel-Softmax, effectively aligning with the softmax function. 
Suppose $S^j_i$ is the score of the $j$-th frame in $\bmV_i$ (generated by Grade-Net), then its Gumbel-Softmax probability is defined by
\begin{gather}
{p_i^j} = \text{Softmax}(S_i^j + \delta \cdot {G_i^j}), \\
{G_i^j} = - \log\left(- \log ({U_i^j}) + \epsilon\right) + \epsilon, \quad {U_i^j} \sim {U}(0,1).
\end{gather} 
Here $\epsilon$ is a small positive constant near zero. $\delta$ is a hyperparameter that controls the noise level in the Gumbel distribution, ensuring different frame sets for the two views. 
Finally, we select and drop $M$ frames with highest scores:
\begin{gather}
\bm{I}_i = \text{ArgTopK}(\{{p_i^1}, {p_i^2}, \dots, {p_i^{M_0}}\},M), \\
\bm{V}_i' = \text{DropIndex}(\bmV_i,\bmI_i)\in \mathbb{R}^{(M_0-M) \times D},
\end{gather}
and remain $M_0-M$ frames for hashing learning.

\subsection{Hashing Network}
\paragraph{Encoder and Decoder}
The encoder and decoder are composed of Vanilla Transformer layers \cite{transformer}.

\paragraph{Hash Layer}
Given the encoded embeddings of the $i$-th video, $\bmZ_i \in \mathbb{R}^{(M_0-M)\times K}$, we obtain the soft hash vector: 
\begin{gather} \label{equ} 
\bm{H}_i = \mathrm{tanh}(\bmZ_i)\in(0,1)^{(M_0-M)\times K},
\end{gather}
where $\mathrm{tanh}$ denotes the hyperbolic tangent function. The video-level hard hash vector is then aggregated via mean pooling and quantization: 
\begin{equation} \label{equ} 
\bm{b}_i = \mathrm{sgn}\left( \text{MeanPool}(\bm{H}_i) \right) \in \{-1, +1\}^K.
\end{equation}
$\mathrm{sgn}$ denotes the sign function. 
The gradient is passed directly through the $\mathrm{sgn}$ function \cite{ste}.

\subsection{Point-to-set Hash-based Contrastive Learning}\label{subsec:Conponent_Voting Hash}

Inspired by DHTA \cite{targetHash}, we propose point-to-set (P2Set) hash-based contrastive learning to align the query's hash code with the hash center of the cluster.

Specifically, we first use the encoder corresponding to the current epoch to encode the training set, obtaining the encoded embeddings, and perform k-means clustering to generate pseudo-labels. Then, based on these pseudo-labels, we apply component voting to compute the hash code center for the videos belonging to the same cluster. 
Finally, by leveraging $\calL_{\text{P2Set}}$, we bring the query closer to the  corresponding cluster center, thereby attracting similar videos and enhancing the neighborhood information of the hash codes.

\paragraph{Component Voting for Hash Centers}
Given a query video $\bmV_q$, the objective is to generate a hash code $\bmb_q$ via our hashing function $\mathcal{H}_t$ such that  $\bmb_q$ is closely aligned with the hash codes $\bmb_{\text{set}}$ of other videos $\bmV_{\text{set}}$ within the same semantic cluster obtained through k-means clustering. This can be formulated as:
\begin{gather}
\min_{\bmV_q} d_H(\mathcal{H}_t(\bmV_q), C(\mathcal{H}_t(\bmV_{\text{set}})))= d_H(\bmb_q, C(\bmb_{\text{set}})),
\end{gather}
where $C(\bmV_{\text{set}})$ denotes the set of videos belonging to the same semantic class, $C(\bmb_{\text{set}})$ refers to the set of hash codes corresponding to videos within the same semantic cluster obtained through k-means clustering. 

To measure the distance between the hash code $\bmb_q$ and the cluster $C(\bmb_\text{set})$, we use the point-to-set metric \cite{targetHash}:
\begin{gather}
d(\bmb_q, C(\bmb_\text{set})) = \frac{1}{|C(\bmb_{\text{set}})|} \sum_{\bmb_{\text{set}} \in C(\bmb_{\text{set}})} d_H(\bmb_q, \bmb_{\text{set}}),
\end{gather}
where $\bmb_q$ is the hash code of the query,  and \(d_H\) is the Hamming distance. The objective is to minimize the distance between the query video and the hash codes of other videos within its corresponding cluster.

We provide a theoretical proof that the aggregated Hamming distance between the query hash codes and the hash center obtained via component voting is equal to $d(\bmb_q, C(\bmb_\text{set}))$, as detailed in the appendix. The hash center is computed via component voting as follows:
\begin{equation}
N_{+1}^j = \sum_{i=1}^{|C(\bmb_{\text{set}})|} \mathbb{I}\left(\bmb_{i}^j = +1 \right),
\end{equation}
\begin{equation}
N_{-1}^j = \sum_{i=1}^{|C(\bmb_{\text{set}})|} \mathbb{I}\left(\bmb_{i}^j = -1 \right),
\end{equation}
\begin{equation}
\bmb_c^j = \begin{cases}
+1, & \text{if } N_{+1}^j \geqslant N_{-1}^j,\\
-1, & \text{otherwise},
\end{cases}
\end{equation}
where $\mathbb{I}( \cdot )$ is an indicator function, and $\bmb_c^j$ represents the value of the $j$-th bit in the hash code of the hash center in a $K$-bit hash vector.  Repeating the above process $N_k $ times will yield a $ K $-bit hash center $\bmb_c$.

\subsection{Self-Supervised Learning Tasks}\label{subsection:Loss}

\paragraph{Frame Reconstruction}
Following ConMH \cite{ConMH}, we select frames with reconstruction masks for efficient reconstruction. The Frame Reconstruction Loss targets hard-to-reconstruct frames, increasing task difficulty and improving the model's encoding performance, described as:
\begin{gather}
   \calL_\mathsf{FR} = \frac{1}{NM} \sum_{i=1}^{N} \sum_{m=1}^{M} \left\| \bmv_{i}^{m} - \bm{\hat{v}}_{i}^{m} \right\|_2^2 ,
\end{gather}
where $\bm{v}_i^m$ denotes the feature of the $m$-th frame of the original input for the $i$-th video, and $\bm{\hat{v}}_i^m$ represents the reconstructed features corresponding to the respective frames. 

\paragraph{View Contrastive Learning}
We align video-level hash codes across different views through view contrastive learning, where two sequences extracted from the same video are treated as positive sample pairs, while sequences from different videos serve as negative sample pairs, namely: 
\begin{equation}
 \calL_\mathsf{VC}^{(i,j)}= - \log \frac{e^{\cos(\bmb_i, \bmb_j) / \tau_{1}}}{ e^{\cos(\bmb_i, \bmb_j) / \tau_{1}} + \sum_{k=1}^{2N} e^{\cos(\bmb_i,\bmb_k)/\tau_{1}}  },
\end{equation}
\begin{equation}
  \calL_\mathsf{VC}= - \frac{1}{2N}\sum_{i=1}^{N}(\calL_\mathsf{VC}^{(i,2i)} +  \calL_\mathsf{VC}^{(2i,i)})  ,
\end{equation}
where $\bmb_i$ and  $\bmb_{2i}$ are positive sample pairs, $\bmb_k$ is considered a negative sample with respect to $\bmb_i$ and $\bmb_j$, and $ \tau_{1}$ $>$ 0.

\paragraph{Point-to-set Hash-based Learning}
Component voting is used to obtain the anchor hash code, followed by point-to-set (P2Set) hash contrastive learning between hash codes from different views and the hash centers, promoting higher-level semantic learning. We compute the P2Set loss as:
\begin{gather}
\calL_\mathsf{P2Set}= - \sum_{k=1}^{N_K}\sum_{i=1}^{2N}\log  \frac{e^{\cos(\bmb_i, \bmb_k^c) / \tau_2}}{  \sum_{m=1}^{N^a_k} e^{\cos(\bmb_i,\bmb_k^m)/\tau_2}  }.
\end{gather}
$N_k$ denotes the number of clustering iterations performed with different numbers of cluster centers.
$\bmb_i$ denotes the hash code of the \(i\)-th video from one view, and $\bmb_k^c$ refers to the belonging cluster center of the \(i\)-th video. $N^a_k$ represents the number of anchors. $\tau_2$ is the temperature factor.

 \paragraph{Aggregate Loss}\label{subsub:Aggregate Loss}
\begin{gather}\label{Aggress Loss}
    \calL_\mathsf{\modelname{}} = \calL_\mathsf{FR}+\alpha \calL_\mathsf{VC}+\beta \calL_\mathsf{P2Set},
\end{gather}
where $\alpha$  and $\beta$ are hyper-parameters to balance loss terms.

\subsection{Adversarial Hashing Learning}

\modelname{} consists of two main components: the Adversarial Automated Sampler and the Hash Network, with a non-parametric Gradient Reversal Layer (GRL) in between.  Given the input full view embeddings of the video $\bm{V} \in \mathbb{R}^{N \times M \times D}$, we consider the function:
\begin{equation} 
\label{eq:min-max}
\calL_{\mathsf{\modelname{}}}= \calL_{\mathsf{\modelname{}}}\left(\mathcal{H}_t\left(\mathcal{S}_t\left(\bmV ; \theta_{\mathcal{S}_t}\right) ; \theta_{\mathcal{H}_t}\right)\right),
\end{equation}
where $\calL_{\mathsf{\modelname{}}}$ represents the total loss, $\mathcal{S}_t$ denotes the sampler, and $\mathcal{H}_t$ refers to the Hashing Network, with their corresponding parameters being $\theta_{\mathcal{S}_t}$ and $\theta_{\mathcal{H}_t}$, respectively. 

\paragraph{Gradient Reversal Layer}
To facilitate single-stage adversarial training, a non-parametric Gradient Reversal Layer (GRL) is incorporated at the end of the sampler, acting as an identity operation during forward propagation and reversing the gradient by multiplying it by -1 during backpropagation.

We compute parameter updates using the SGD algorithm:
\begin{equation}
\label{eq:update}
\begin{aligned}
    \theta_{\mathcal{S}_t}  &\longleftarrow  \theta_{\mathcal{S}_t}+\mu \frac{\partial \calL_{\mathsf{\modelname{}}}}{\partial \theta_{\mathcal{S}_t}}, \\
\theta_{\mathcal{H}_t}  &\longleftarrow  \theta_{\mathcal{H}_t}-\mu \frac{\partial \calL_{\mathsf{\modelname{}}}}{\partial \theta_{\mathcal{H}_t}},\\
\end{aligned}
\end{equation}
where $\mu$ is the learning rate. Due to the presence of the GRL, $\theta$ is ultimately updated through gradient ascent.

Based on \Cref{eq:update}, we are actually seeking the parameters $\hat{\theta}_{\mathcal{S}_t}$, $\hat{\theta}_{\mathcal{H}_t}$
that deliver a saddle point of \Cref{eq:min-max}:
\begin{equation}
\begin{gathered}
\label{eq:final}
\hat{\theta}_{\mathcal{H}_t}=\arg \min _{ \theta_{\mathcal{H}_t}} \calL_{\mathsf{\modelname{}}}\left(\hat{\theta}_{\mathcal{S}_t}, \theta_{\mathcal{H}_t}\right), \\
\hat{\theta}_{\mathcal{S}_t}=\arg \max _{\theta_{\mathcal{S}_t}} \calL_{\mathsf{\modelname{}}}\left(\theta_{\mathcal{S}_t}, \hat{\theta}_{\mathcal{H}_t}\right).
\end{gathered}
\end{equation}

As seen from \Cref{eq:final}, while the sampler aims to maximize the all loss, the hashing network concurrently seeks to minimize it. These two components participate in a min-max game, where they counterbalance and co-evolve. The three distinct loss components each play distinct roles in the sampling process: $\calL_{\text{FR}}$ ensures the adaptive sampling of frames that are difficult to reconstruct, $\calL_{\text{VC}}$ facilitates the adaptive sampling of frame sequences from two different views that are as dissimilar as possible, serving as a data augmentation technique to improve the effectiveness of VC, and $\calL_{\text{P2Set}}$ adaptively samples frame sequences that are somewhat distant from the center,  enhancing the model's robustness.
Ultimately, the sampler adaptively selects frames that are more challenging to reconstruct, while the Hashing Network concurrently refines its encoding capacity. This dynamic interaction enhances hash code retrieval effectiveness.

\section{Experiments} \label{sec:experiments}

\begin{figure*}[t] 
\centering
{\includegraphics[width = 0.9\textwidth]{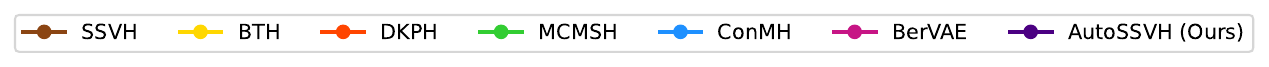}} 
\vspace{-1em}
  \setcounter{subfigure}{0}

\subfloat[\small{ActivityNet 16 bits}]{\includegraphics[width = 0.22\textwidth]{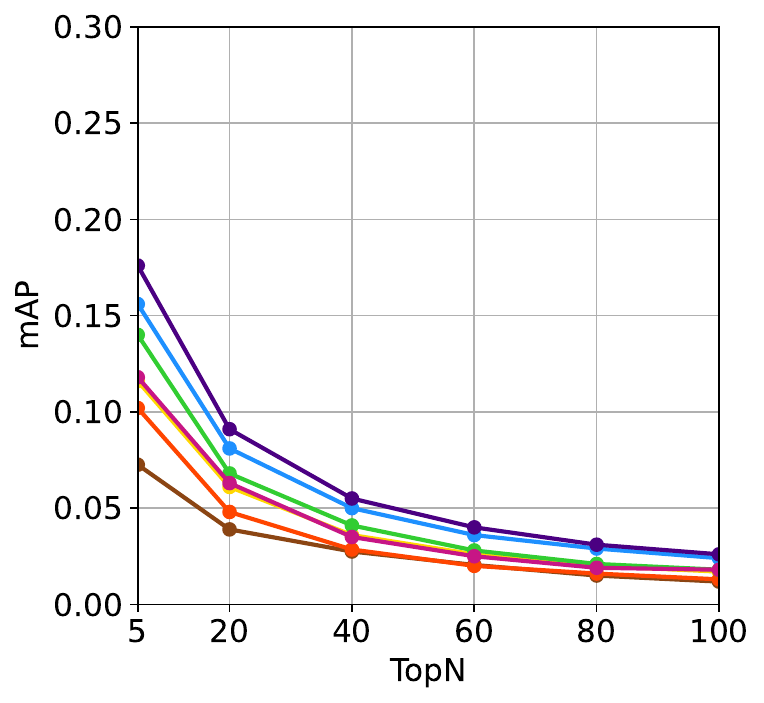}}
  \subfloat[\small{FCVID 16 bits}]{\includegraphics[width = 0.22\textwidth]{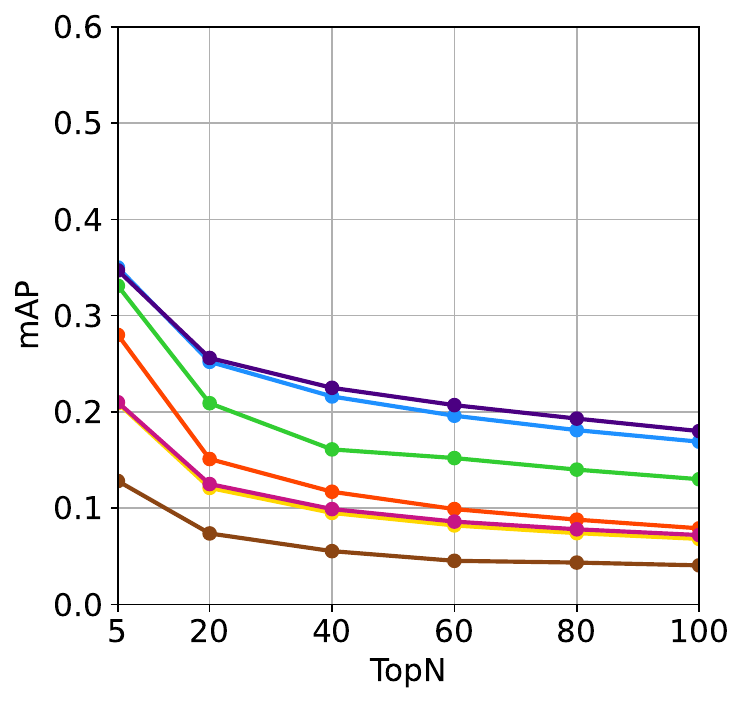}}
    \subfloat[\small{UCF101 16 bits}]{\includegraphics[width = 0.22\textwidth]{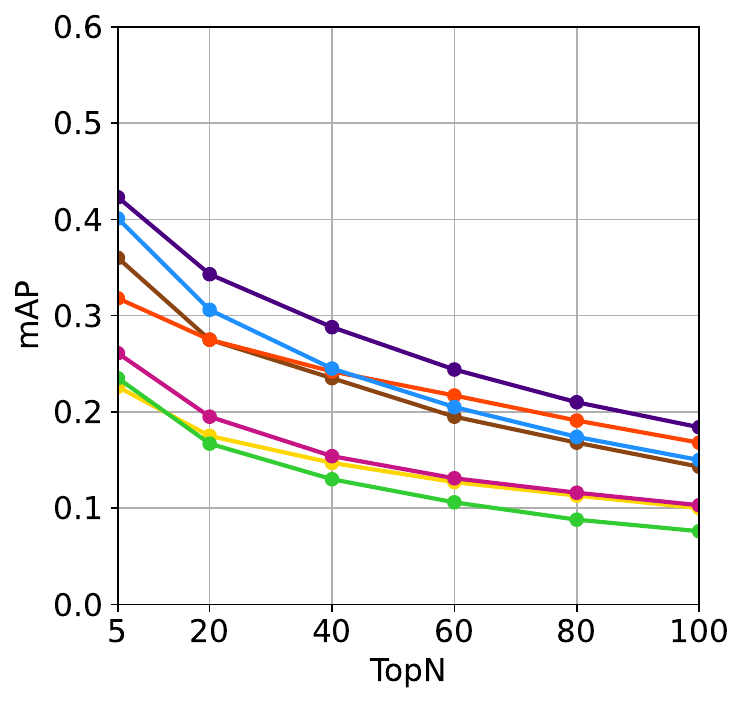}}
      \subfloat[\small{HMDB51 16 bits}]{\includegraphics[width = 0.22\textwidth]{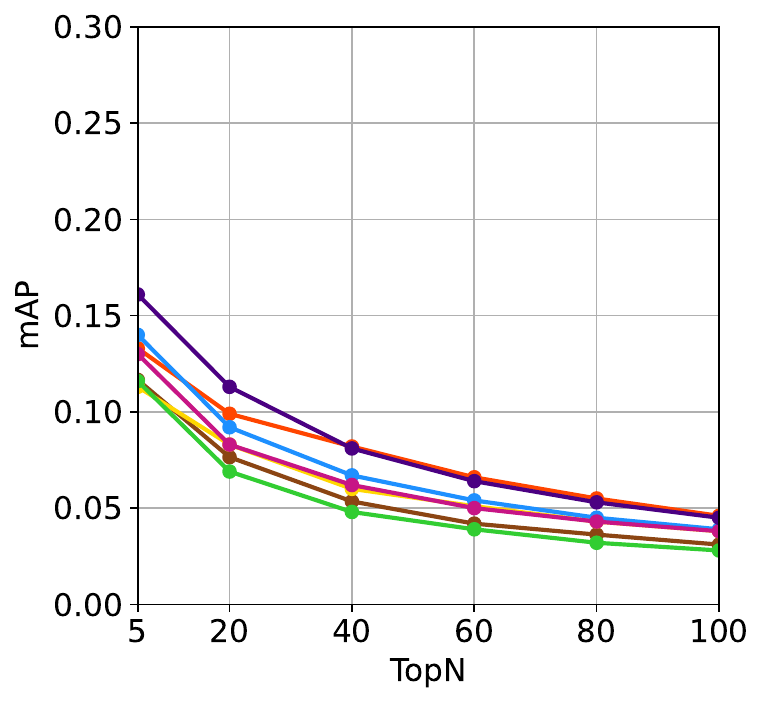}}
\vspace{1em}
\subfloat[\small{ActivityNet 32 bits}]{\includegraphics[width = 0.22\textwidth]{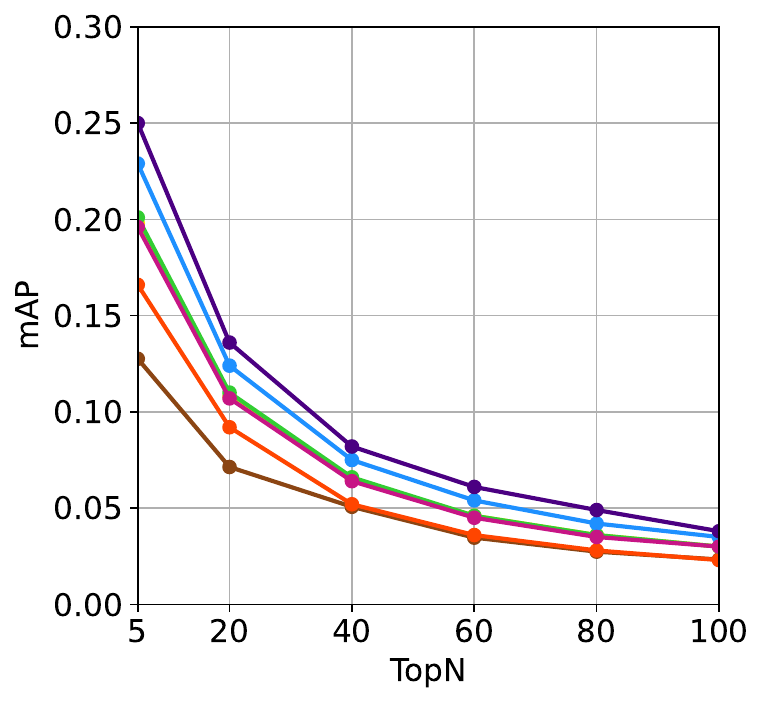}}
  \subfloat[\small{FCVID 32 bits}]{\includegraphics[width = 0.22\textwidth]{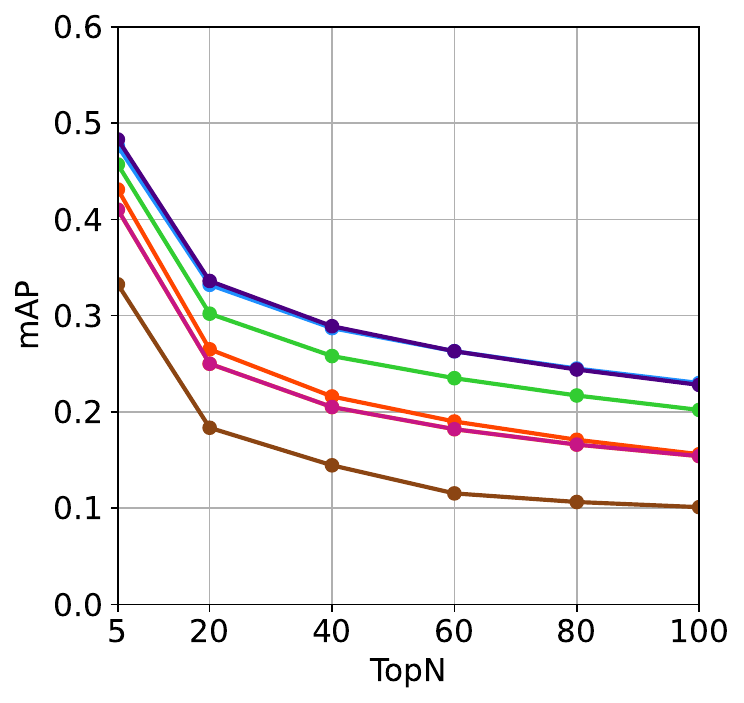}}
    \subfloat[\small{UCF101 32 bits}]{\includegraphics[width = 0.22\textwidth]{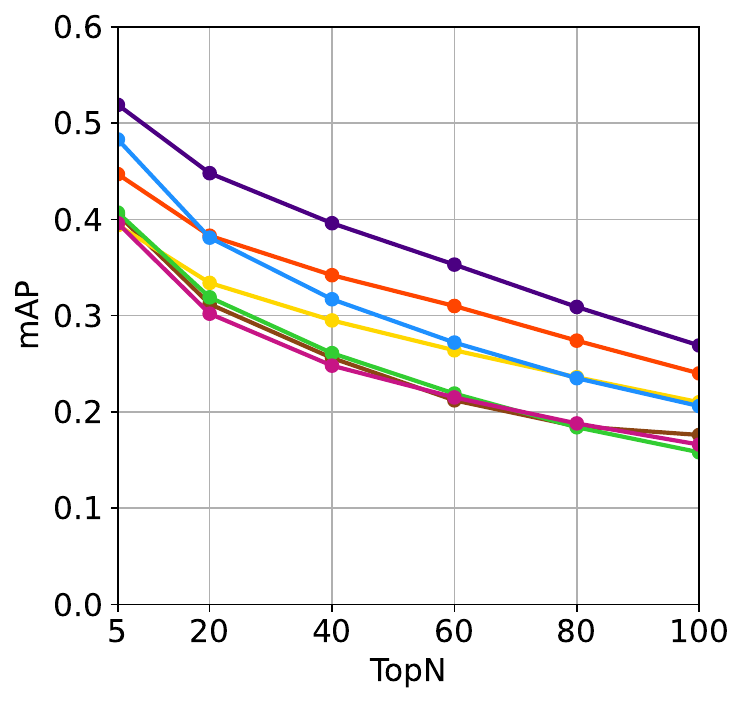}}
      \subfloat[\small{HMDB51 32 bits}]{\includegraphics[width = 0.22\textwidth]{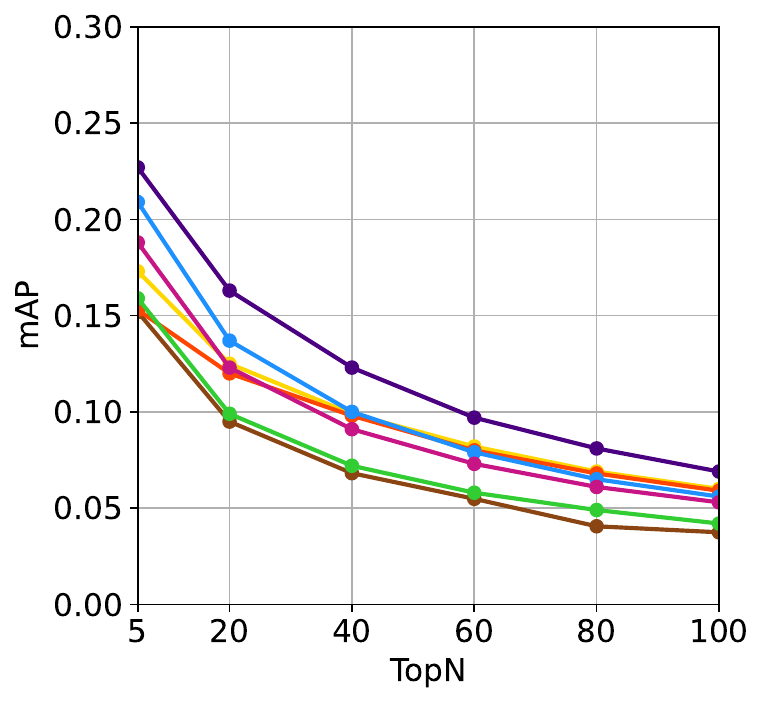}}
      \vspace{0em}
\subfloat[\small{ActivityNet 64 bits}]{\includegraphics[width = 0.22\textwidth]{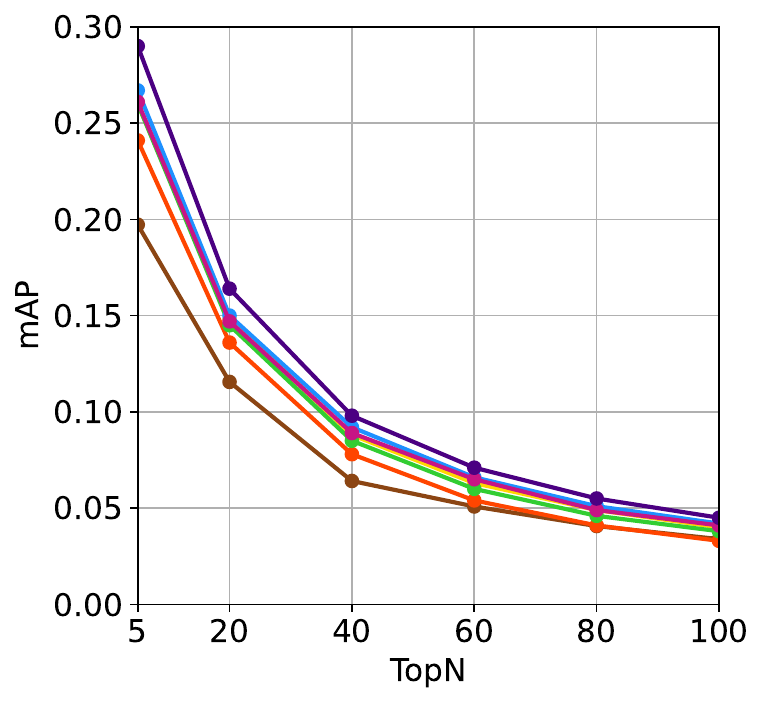}}
  \subfloat[\small{FCVID 64 bits}]{\includegraphics[width = 0.22\textwidth]{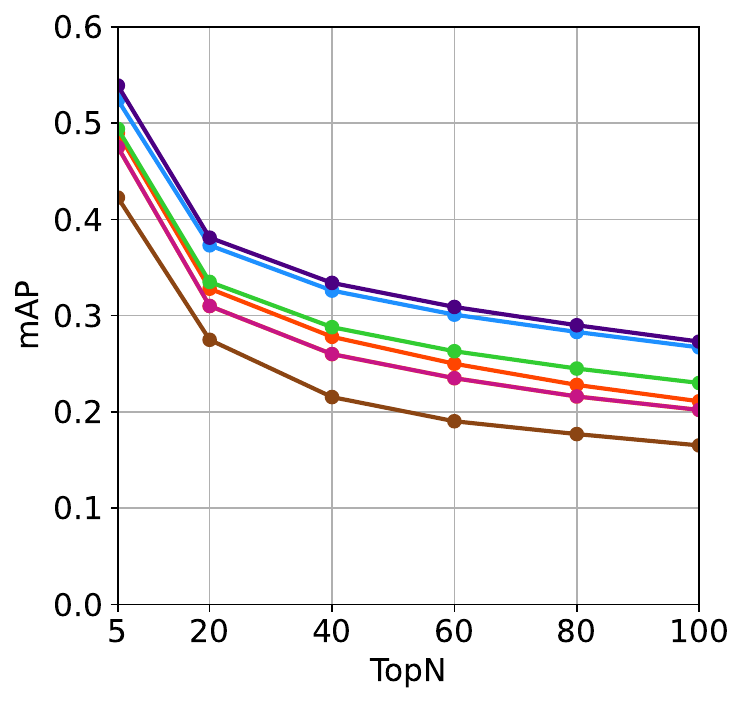}}
    \subfloat[\small{UCF101 64 bits}]{\includegraphics[width = 0.22\textwidth]{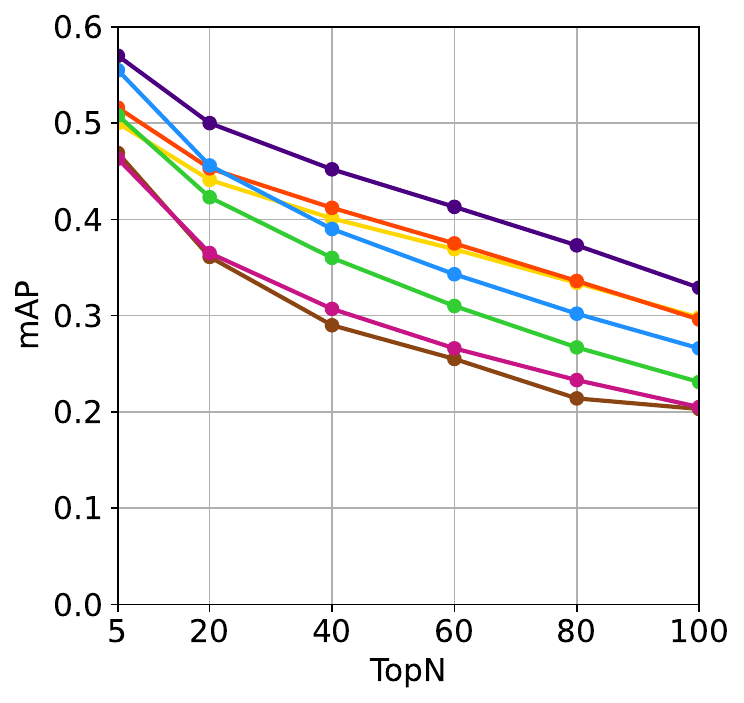}}
  \subfloat[\small{HMDB51 64 bits}]{\includegraphics[width = 0.22\textwidth]{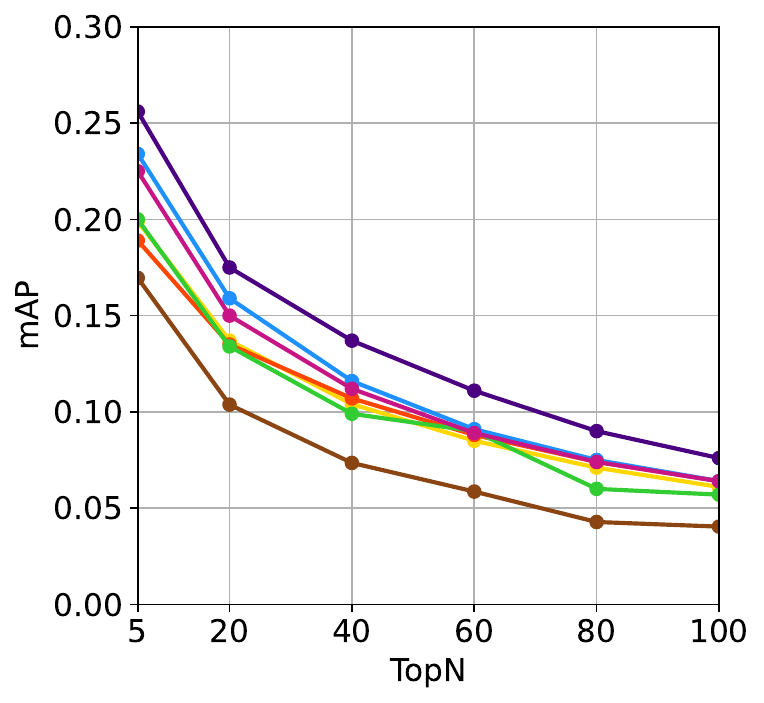}}
  \caption{Comparison of retrieval performance using mAP@N on  ActivityNet, FCVID, UCF101 and HMDB51. }
\label{fig:map}
\end{figure*}

\subsection{Dataset}\label{subsec:dataset}

To ensure a fair comparison with current SOTA methods, we selected four benchmark datasets widely used in the field of self-supervised video hashing: \textbf{ActivityNet \cite{Activitynet}, FCVID \cite{FCVID}, UCF101 \cite{UCF} and HMDB51 \cite{HMDB}}.
\textbf{ActivityNet}  contains approximately 20,000 YouTube videos distributed across 200 distinct activity categories. For training, we selected a subset of 9,722 videos. Due to the unavailability of the official test partition, we repurposed the validation set as the test set. Within this set, 1,000 videos were randomly chosen as query instances, while the remaining 3,760 videos were designated as the retrieval database. \textbf{FCVID}  comprises 91,223 web videos, manually labeled into 239 categories. After filtering out corrupted data and resolving category overlaps, we curated a subset of 91,185 videos. Consistent with  ConMH \cite{ConMH}, we allocated 45,585 videos for training, while the remaining 45,600 videos were divided for query and retrieval tasks.
 \textbf{UCF101} consists of 13,320 videos, covering 101 categories of human actions. We adhered to the CHAIN \cite{CHAIN} protocol by using 9,537 videos for training and retrieval, with the remaining 3,783 videos from the test set serving as queries.
 \textbf{HMDB51}  contains 6,849 videos across 51 action categories. Following  CHAIN , we employed 3,570 videos for training and retrieval, with the test set providing 1,530 videos for query purposes.

\subsection{Implementation Details}\label{subsection:Implementation Details}

\paragraph{Data Pre-processing}
 We extract frame embeddings from videos using models pre-trained on ImageNet \cite{russakovsky2015imagenet}: VGG-16 \cite{vgg} for ActivityNet (30 frames, 1024-dimensional) and ResNet-50 \cite{ResNet} for other datasets (25 frames, 2048-dimensional), ensuring consistency with ConMH \cite{ConMH}.

\paragraph{Model Architecture}
In the case of ActivityNet, \modelname{} employs an encoder and decoder with 6 and 1 Transformer layers, respectively, while 12 and 2 layers are adopted for all other datasets. The hidden layer size ratio is set to 4.

\paragraph{Training Details}
For ActivityNet and FCVID, we set $\mathbf{\alpha}$ to 0.2, $\mathbf{\beta}$ to 0.01, and the warm-up period to 100 epochs. For HMDB51 and UCF101, $\mathbf{\alpha}$ is set to 1, $\mathbf{\beta}$ to 0.2, and the warm-up period to 50 epochs. The number of clustering iterations is set to three, with cluster center sizes of 250, 400, and 600, respectively. Adam \cite{kingma2014adam} is used as the optimizer, and further experimental details can be found in the appendix.
Our training procedure is conducted in two phases. The first phase serves as a warm-up, where the model is trained without the component voting mechanism, allowing it to capture lower-level semantic information. Following a number of epochs, the second phase is initiated, introducing component voting to facilitate the model's ability to capture higher-level semantic information,  accelerating convergence.

\paragraph{Evaluation Protocols}
Following prior work \cite{ConMH}, we assess performance using mean Average Precision at top-$N$ results (\textbf{mAP@$N$}), with $N \in {5, 20, 40, 60, 80, 100}$. To enable a more detailed evaluation, we also report \textbf{Precision-Recall (PR)} curves, which illustrate performance across varying decision thresholds. Additionally, we introduce an overall metric to provide a holistic summary,
\begin{equation}
\textbf{GMAP}=\sqrt{\sum_{N \in \{5,20,40,60,80,100\}}(\mathrm{mAP} @ N)^2},
\end{equation}
which computes the geometric mean of mAP results, providing a comprehensive assessment across retrieval thresholds.

\subsection{Comparison with State-of-the-arts}  \label{subsec:sota_experiments}

\paragraph{Baselines}
We selected a set of widely recognized and open-source baselines in the field of self-supervised video hashing: SSVH \cite{SSVH}, BTH \cite{BTH}, DKPH \cite{DKPH}, MCMSH \cite{MCMSH}, ConMH \cite{ConMH}, and BerVAE \cite{BerVAE}.  All methods were trained and evaluated under consistent experimental conditions.

\paragraph{MAP Comparison}
As illustrated in  \Cref{fig:map}, \modelname{} consistently outperforms all baselines across all bit lengths on all datasets, establishing a new state-of-the-art. This performance is largely attributed to the contributions of our adversarial automated sampling module and the synergistic effects of  component voting hash learning. Specifically, on UCF101 and HMDB51, \modelname{} exceeds the best competitor, ConMH, for 16-bit, 32-bit, and 64-bit representations, respectively. On ActivityNet and FCVID, significant improvements in GMAP are also observed, with relative gains for 16-bit, 32-bit, and 64-bit hash lengths, respectively. These results demonstrate the generalizability of \modelname{}, achieving outstanding performance across diverse  datasets.

\begin{table}[t]
\centering
\resizebox{\columnwidth}{!}{
    \begin{tabular}{cccccccc}
    \toprule
     \multirow{2}{*}{Method} 
    & \multicolumn{3}{c}{UCF101}  &  
    & \multicolumn{3}{c}{HMDB51}  \\
     & N=60& N=80& N=100& & N=60& N=80& N=100\\
    \midrule
 ConMH& $\uparrow10.6\%$& $\uparrow13.1\%$& $\uparrow15.2\%$& &$\uparrow1.1\%$& $\uparrow25.0\%$& $\uparrow12.3\%$\\  \midrule\midrule
\textbf{AutoSSVH}&          {$\uparrow \textbf{36.5}\%$}&     {$\uparrow \textbf{43.4\%}$}&          {$\uparrow \textbf{46.7} \%$}&& \textbf{$\uparrow \textbf{34.4} \%$}& \textbf{$\uparrow \textbf{66.7} \%$}& \textbf{$\uparrow \textbf{50.9}\%$}\\ \midrule
    \end{tabular}}
     \vspace{-.5em}
    \caption{Relative Percentage Improvements in GMAP of AutoSSVH and ConMH over MCMSH at Higher $N$  on UCF101 and HMDB51 with 64-bit hash codes.}
    \label{tab:global features comparison}
    \vspace{-1em}
\end{table}

Observing  \Cref{fig:map}$(g)–(l)$, it is evident that \modelname{} shows significant improvements, particularly at higher values of $N$ in map@N. As demonstrated by \Cref{tab:global features comparison}, 
 \modelname{} achieves  great improvements over MCMSH, with increases of $\mathbf{34.4\%}$, $\mathbf{66.7\%}$, and $\mathbf{50.9\%}$ for the same $N$ values. A similar trend is observed on  UCF101, where \modelname{} continues to outperform ConMH. These results are attributed to the effectiveness of our component voting strategy in capturing complex global information and high-level semantics.

\begin{figure}[t]
\centering
{\includegraphics[width = 0.48\textwidth]{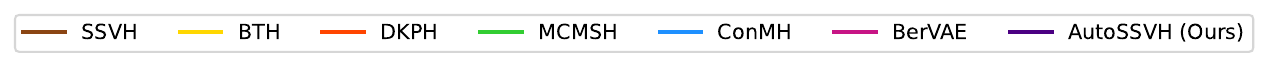}}
    \setcounter{subfigure}{0}
  \subfloat[\small{UCF101 16 bits}]{\includegraphics[width = 0.1500\textwidth]{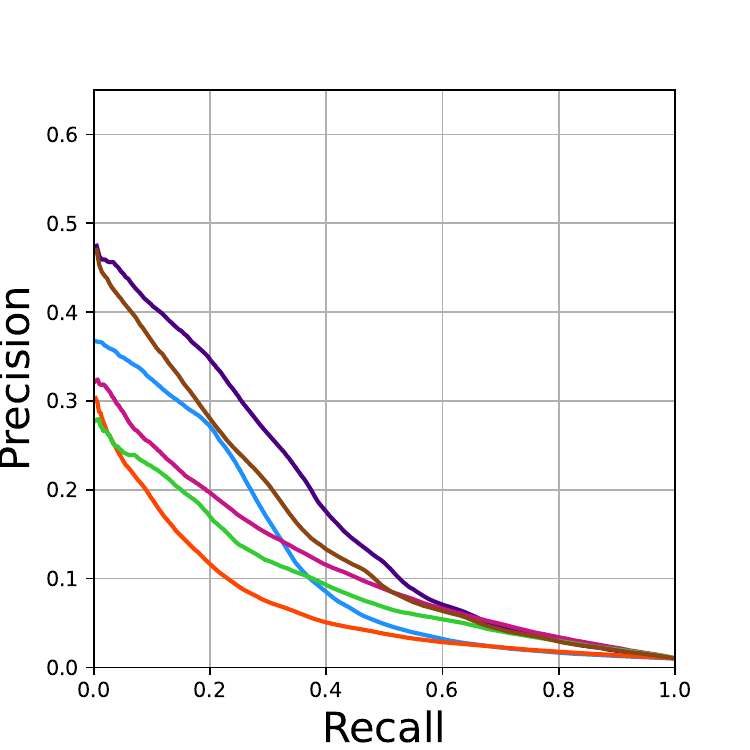}}
  \subfloat[\small{UCF101 32 bits}]{\includegraphics[width = 0.1500\textwidth]{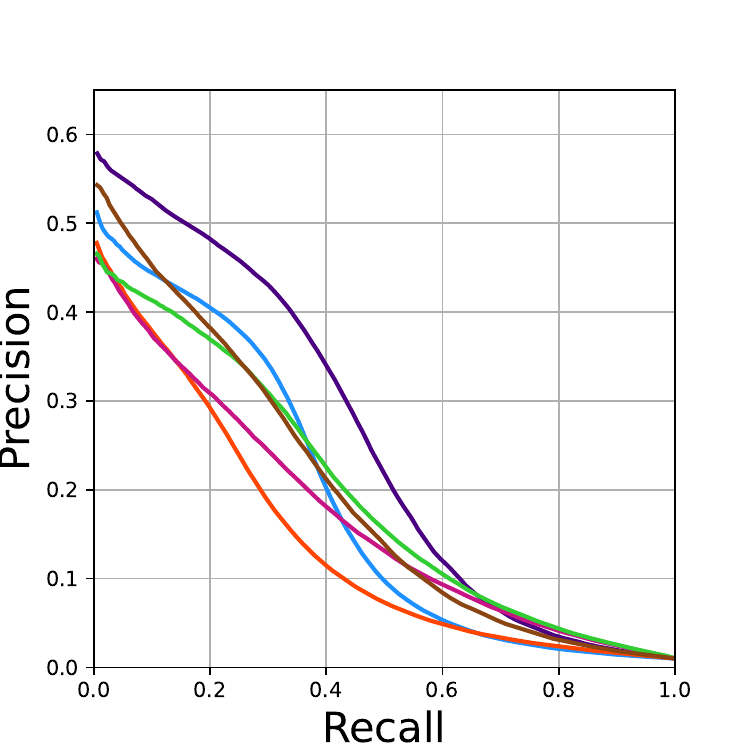}}
  \subfloat[\small{UCF101 64 bits}]{\includegraphics[width = 0.1500\textwidth]{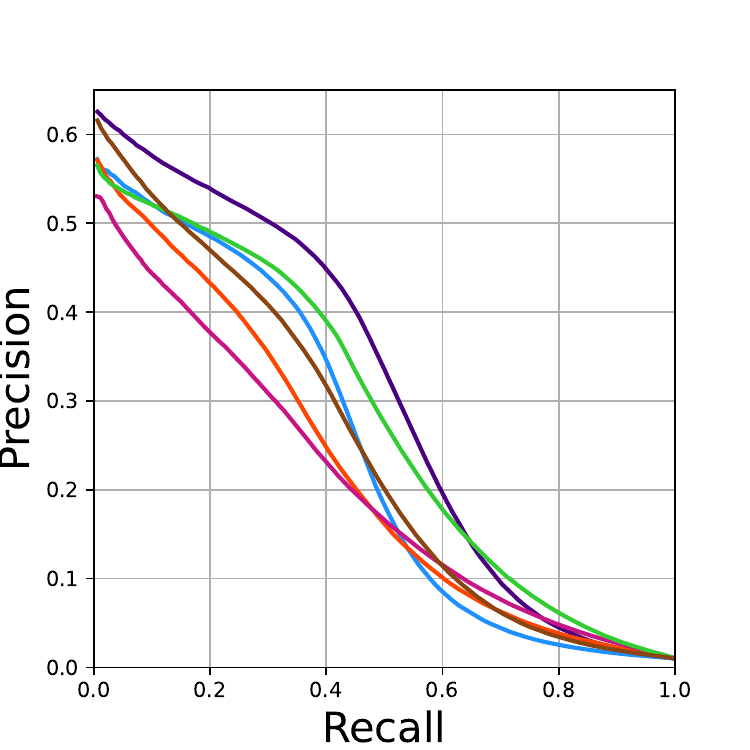}}
  \caption{Retrieval PR curves of different models on  UCF101.}
  \label{fig:PR_curve}
\end{figure}

\paragraph{PR Curve Analysis}
To assess the retrieval performance across a broader range of ranking positions, we present the PR curves for various models. As depicted in the  \Cref{fig:PR_curve}, 
our approach consistently outperforms other state-of-the-art methods, achieving higher precision and recall, with both metrics reaching the outer ranking positions more effectively across all evaluated hash code lengths.

\begin{table}[t]
\centering
\resizebox{\columnwidth}{!}{
    \begin{tabular}{ccccc}
    \toprule
     {Method}& 16 bits & 32 bits & 64 bits & Per Impr\\\midrule
BerVAE&        0.3350&      0.0481& 0.0648& $\uparrow$4.1\%\\
 ConMH& 0.0285& 0.0482& 0.0640&0.0\%\\
 MCMSH& 0.0229& 0.0461& 0.0582&$\downarrow$9.6\%\\
 DKPH& 0.0243& 0.0452& 0.0550&$\downarrow$11.5\%\\
 BTH& 0.0210& 0.0418& 0.0525&$\downarrow$18.1\%\\ 
\midrule\midrule
 \textbf{\modelname{}(Ours)}& \textbf{0.0410}
& \textbf{0.0550}& \textbf{0.0780} &$\uparrow$ \textbf{20.7}\%\\\midrule
    \end{tabular}}
    \caption{ Cross-dataset retrieval performance, measured by GMAP, is evaluated by training on UCF101 and testing on HMDB51. }
    \label{tab:cross dataset validation}

\end{table}

\paragraph{Cross-dataset Validation}
To assess the impact of adversarial training on the model's generalization and transferability, we conducted experiments training on UCF101 and validating on HMDB51 with varying bit-widths. As shown in \Cref{tab:cross dataset validation}, our model demonstrates improvements across all bits. Additionally, we computed the geometric mean of the percentage improvement over ConMH. The results indicate that \modelname{} maintains, and even enhances, its generalization ability, showing a $\mathbf{20.7\%}$ improvement.

\subsection{Model Analyses}  \label{subsec:ablation_study}

\subsubsection{Ablation Study Analysis}\label{subsubsec:Abaltion}

We conducted ablation studies on the UCF101 and HMDB51 datasets to evaluate the impact of the key contributions of \modelname{} on its overall performance.

\begin{table}[t]
\centering
\resizebox{\columnwidth}{!}{
    \begin{tabular}{ccccccccc}
    \toprule
    \multirow{2}{*}{ID} & \multirow{2}{*}{Method} 
    & \multicolumn{3}{c}{UCF101}  &  
    & \multicolumn{3}{c}{HMDB51}  \\
    \cmidrule{3-5} \cmidrule{7-9}
    & & 16 bits & 32 bits & 64 bits & & 16 bits & 32 bits & 64 bits \\  \midrule\midrule
(I)&w/o $\mathcal{ADV}$&          0.699&     0.941&          0.987&& 0.217& 0.312& 0.351\\
(II)&w/ $Random$&        0.700&      0.945& 0.988&& 0.213& 0.316& 0.355\\
 (III)& w/ $AttMask$& 0.701& 0.942& 0.991& & 0.219& 0.324&0.366\\
 (IV)& w/ $AdaMAE$& 0.705& 0.948& 0.995& & 0.223& 0.321&0.364\\
 (V)& w/ $ADIOS$& 0.704& 0.947& 0.997& & 0.225& 0.328&0.369\\ \midrule\midrule
(VI)& w/o $\calL_\mathsf{FR}$& 0.709& 0.945& 1.01&& 0.227& 0.325& 0.359\\
(VII)& w/o $\calL_\mathsf{VC}$& 0.701& 0.942& 0.992&& 0.219& 0.321& 0.360\\
(VIII)& w/o $\calL_\mathsf{P2Set}$& 0.705& 0.948& 0.990&& 0.214& 0.322& 0.365\\ \midrule\midrule
 (IX)& \textbf{\modelname{} (full)}& \textbf{0.719}& \textbf{0.959}& \textbf{1.09}& & \textbf{0.233}& \textbf{0.338}&\textbf{0.376}\\\midrule
    \end{tabular}}
    \vspace{-.5em}
    \caption{Ablation studies of \modelname{} evaluated using GMAP.}
    \label{tab:ablation study}
    \vspace{-1em}
\end{table}

\paragraph{Effectiveness of $\mathcal{ADV}$}
In this setting, we remove the gradient reversal layer (GRL). As illustrated in the row (I) of \Cref{tab:ablation study}, for all bit sizes across both UCF101 and HMDB51, the removal of the GRL leads to an approximate \textbf{5.9\%} decrease in GMAP accuracy. This finding underscores the critical role of the gradient reversal layer in enhancing retrieval performance by facilitating the learning of more discriminative features. Its absence impairs the model's capacity to preserve the semantic integrity of the generated hash codes.
 
 \paragraph{Effectiveness of Sampler Strategy}
 The w/ $Random$ strategy employs a random masking approach, and we also conducted supplementary experiments with three commonly used sampling strategies: AttMask \cite{attmask}, AdaMAE \cite{bandara2023adamae}, and ADIOS \cite{adios}. As observed in  \Cref{tab:ablation study}(II) - (V), our sampling strategy outperforms other sampling methods by an average of \textbf{4.7\%}. Our strategy is not only more lightweight but also yields superior performance in terms of effectiveness.

\paragraph{Effectiveness of $\calL_\mathsf{FR}$}
$\calL_{\text{FR}}$ is responsible for reconstructing the video, which directly influences the model's ability to reconstruct and, consequently, the semantic integrity of the hash codes. As depicted in  \Cref{tab:ablation study}(VI) , the ablation of FR results in a decrease of approximately \textbf{3.9\% }in GMAP.

\paragraph{Effectiveness of $\calL_\mathsf{VC}$}
VC is essential for learning low-level semantic information by modeling different views of a single video, thus ensuring that the hash codes reflect the neighborhood information of lower-level semantics. As shown in  \Cref{tab:ablation study}(VII), the VC loss contributes approximately an \textbf{5.1\%} increase in GMAP accuracy.

\paragraph{Effectiveness of  $\calL_\mathsf{P2Set}$}
P2Set hash learning component is crucial for learning global semantic representations, while its backpropagation mechanism assists the automated sampling module in selecting frames that are more consistent with global semantics. This significantly increases the neighborhood information captured in the generated hash codes, leading to improved retrieval accuracy. As detailed in   \Cref{tab:ablation study}(VIII), the removal of P2Set hash-based learning results in a reduction of GMAP by about \textbf{4.8\%}.

\begin{figure}[t]
\centering
    \vspace{-.5em}
    \setcounter{subfigure}{0}
\includegraphics[width = 0.4500\textwidth]{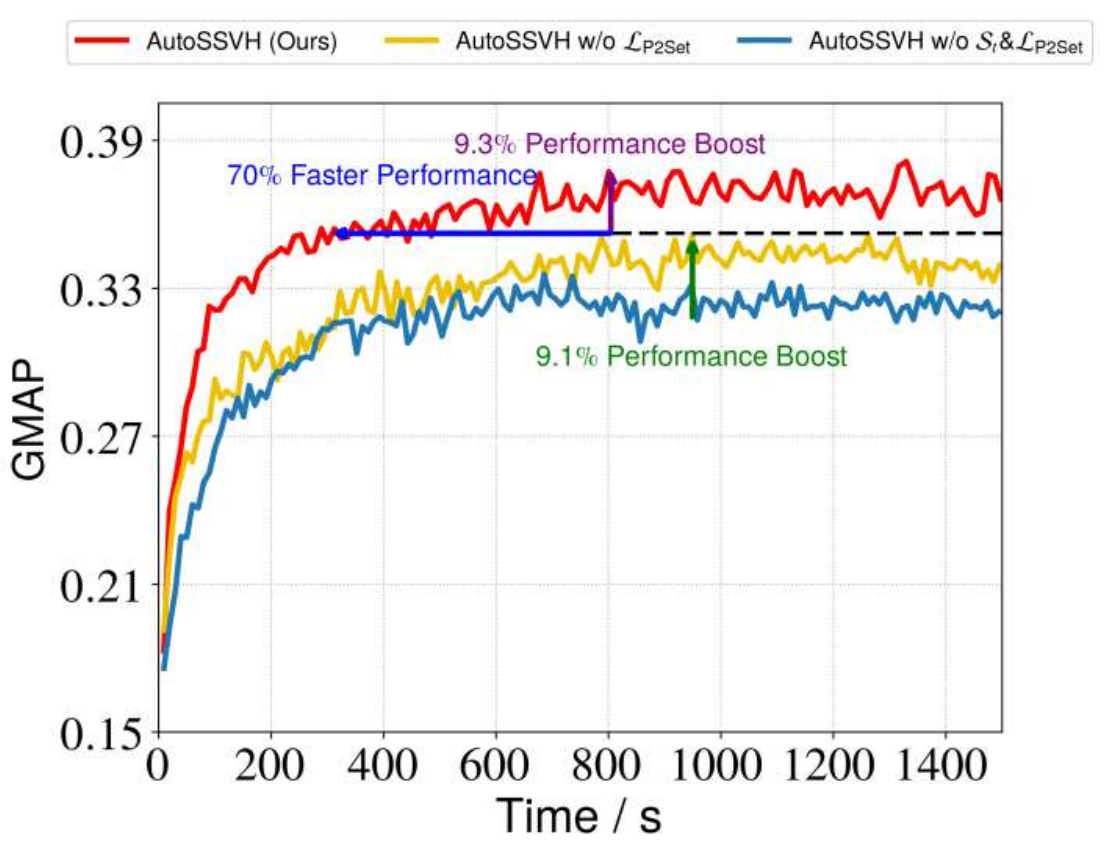}
  \caption{The impact of the automated adversarial sampling strategy and point-to-set (P2Set) hash-based learning on the retrieval efficiency  of \modelname{}. }
  \label{fig:comparative analysis}
\end{figure}

\vspace{-1em}
\subsubsection{Comparative Analysis of Efficiency}\label{subsubsec:Efficiency and Effectiveness}

To further investigate the efficiency and effectiveness of \modelname{}, we conducted an efficiency analysis experiment. As shown in  \Cref{fig:comparative analysis}, our adversarial module achieved a relative improvement of 9.1\%, which further demonstrates the effectiveness. Additionally, the inclusion of P2Set hash learning not only led to a 9.3\% improvement but also resulted in a 70\% speedup, aligning with our expectations and validating the high efficiency and performance.

\section{Conclusions}
\label{sec:conclusion}

In this paper, we introduce \modelname{}, a novel self-supervised video hashing framework that employs adversarial strategies for automated hard-frame sampling. To accelerate the convergence of adversarial training, we incorporate a component voting hash mechanism, facilitating a synergistic integration of the two approaches for the efficient generation of hash codes that capture enriched semantic information and refined neighborhood relationships. Extensive experimental evaluations on four widely adopted benchmark datasets demonstrate that \modelname{} outperforms existing state-of-the-art methods, offering both rapid and effective video retrieval. Our study underscores the strong potential of adversarial strategy-based automated frame sampling for video hashing, which we hope will inspire future researches.

\paragraph{Acknowledgments}
We thank the anonymous reviewers and chairs for their efforts and constructive suggestions. 
This work is supported in part by the National Natural Science Foundation of China under grant 624B2088, 62171248, 62301189, the PCNL KEY project (PCL2023AS6-1), and Shenzhen Science and Technology Program under Grant KJZD20240903103702004, JCYJ20220818101012025, RCBS20221008093124061, GXWD20220811172936001.

 \small \bibliographystyle{ieeenat_fullname} \bibliography{main}

\clearpage
\appendix

\section{Further Details of the Method }
\label{sec:Further Details of the Method }
\noindent \subsection{The Process of Component Voting}\label{subsec:The process of component voting}

To further establish the theoretical validity of our component voting mechanism, we present a detailed proof in this section, following the approach outlined in DHTA\cite{targetHash}.
\begin{equation}
\label{eq:update}
\begin{aligned}
\min_{\bmV_q} d_H(\mathcal{H}_t(\bmV_q), C(\mathcal{H}_t(\bmV_{\text{set}}))) = \min_{\bmb_q} d_H(\bmb_q, C(\bmb_{\text{set}})),
\end{aligned}
\end{equation}
where $\bmV_q$ represents the query video and $\bmb_q$ is the hash code linked to $\bmV_q$. $C(\bmV_{\text{set}}))$ denotes the set of videos belonging to the same semantic cluster, $C(\bmb_{\text{set}})$ refers to the set of corresponding hash codes.
\begin{equation}
\label{eq:update}
\begin{aligned}
\overline{d_H}(\bmb_q, \bmb_c) = \frac{1}{|C(\bmV_{\text{set}})|} \sum_{\bmV_{\text{set}} \in C(\bmV_{\text{set}})} d_H(\bmb_q, \bmb_{\text{set}}),
\end{aligned}
\end{equation}
where $\bmb_c$ is the set of hash codes in the target cluster, and \(d_H\) is the Hamming distance.
\begin{equation}
\label{eq:update}
\begin{aligned}
\min_{\bmb_q} \frac{1}{|C(\bmb_{\text{set}})|} \sum_{\bmb_{\text{set}} \in C(\bmb_{\text{set}})} d_H(\bmb_q, \bmb_{\text{set}}),
\end{aligned}
\end{equation}
\begin{equation}
\label{eq:update}
\begin{aligned}
\bmb_c =\argmin{\bmb_{c'} \in \{+1, -1\}^K} \sum_{i = 1}^{|C(\bmV_{\text{set}})|} d_H(\bmb_{c'},\bmb_i).
\end{aligned}
\end{equation}
We adopt the average-case metric, and as such, the aforementioned constitutes our ultimate optimization objective.
\begin{equation}
N_{+1}^j = \sum_{i=1}^{|C(\bmV_{\text{set}})|} \mathbb{I}\left(\bmb_{i}^j = +1 \right),
\end{equation}
\begin{equation}
N_{-1}^j = \sum_{i=1}^{|C(\bmV_{\text{set}})|} \mathbb{I}\left(\bmb_{i}^j = -1 \right),
\end{equation}
\begin{equation}
\label{eq:voting}
\bmb_c^j = \begin{cases}
+1, & \text{if } N_{+1}^j \geqslant N_{-1}^j, \\
-1, & \text{otherwise}.
\end{cases}
\end{equation}
This outlines the specific process for generating the hash code center $\bmb_c$. Let \( \mathbb{I}( \bullet ) \) denotes an indicator function, and $\bmb_c^j$ represent the value of the $j$-th bit of the hash center $\bmb_c$ within a $K$-bit hash vector. Repeating this process $K $ times results in a $K$-bit hash center $\bmb_c$.

\paragraph{Mathematical proof}
We need to prove that for any $\bmb_{c'} \in \{+1, -1\}^k $, where $\bmb_c \neq \bmb_c'$ , the following inequality holds universally.
\begin{equation}
\label{eq:update}
\begin{aligned}
\sum_{i = 1}^{|C(\bmV_{\text{set}})|} d_H(\bmb_{c},\bmb_i) \leq \sum_{i = 1}^{|C(\bmV_{\text{set}})|} d_H(\bmb_{c'},\bmb_i).
\end{aligned}
\end{equation}
Since $\bmb_c$ is obtained through the Equation (\ref{eq:voting}), it follows that the following equation holds.
\begin{equation}
\label{eq:update}
\begin{aligned}
\sum_{i=1}^{|C(\bmV_{\text{set}})|} \mathbb{I}\left(\bmb_c^j=\bmb_i^j\right) \geq \sum_{i=1}^{|C(\bmV_{\text{set}})|} \mathbb{I}\left(\bmb_{c'}^j=\bmb_i^j\right),
\end{aligned}
\end{equation}
\begin{equation}
\label{eq:update}
\begin{aligned}
\phi(\bmb_c^j, \bmb^j) = \sum_{i=1}^{|C(\bmV_{set})|} \mathbb{I}\left(\bmb_c^j=\bmb_i^j\right),
\end{aligned}
\end{equation}
\begin{equation}
\label{eq:update}
\begin{aligned}
\phi(\bmb_{c'}^j, \bmb^j) =\sum_{i=1}^{|C(\bmV_{\text{set}})|} \mathbb{I}\left(\bmb_{c'}^j=\bmb_i^j\right).
\end{aligned}
\end{equation}

Let $ \Delta = \{ k \mid \bmb_c^k \neq \bmb_{c'}^k \} $ represent the set of indices $k$ where the bit values of $\bmb_c$ and $\bmb_{c'}$ differ. Let $ \overline{\Delta}$  denote the complement of $\Delta$ within the set $ \{1, 2, \dots, k\}$, i.e.,  $\overline{\Delta} = \{1, 2, \dots, k\} \setminus \Delta$.
\begin{align} &\sum_{i=1}^{|C(\bmV_{\text{set}})|} d_H\left(\bmb_c, \bmb_i\right) \\
=& \sum_{j \in \mathcal{\Delta}}\sum_{i=1}^{|C(\bmV_{\text{set}})|} d_H\left(\bmb_c^j, \bmb_i^j\right)+\sum_{j \in \overline{\Delta}}\sum_{i=1}^{|C(V_{\text{set}})|} d_H\left(\bmb_{c}^j, \bmb_i^j\right) \\
=& \sum_{j \in \Delta}\left(M-\phi(\bmb_c^j, \bmb^j)\right)
+\sum_{j \in \overline{\Delta}}\left(M- \phi(\bmb_{c}^j, \bmb^j)\right) \\
\leq& \sum_{j \in \mathcal{D}}\left(M-\phi(\bmb_{c'}^j, \bmb^j)\right)+\sum_{j \in \overline{\Delta}} \left(M-\phi(\bmb_{c'}^j, \bmb^j)\right) \\
=&\sum_{i=1}^{|C(\bmV_{\text{set}})|} d_H\left(\bmb_{c'}, \bmb_i\right). \end{align}
$M$ is defined as $ M = K \times |C(\bmV_{\text{set}})|$, with \( K \) being a constant and $ |C(\bmV_{\text{set}})| $ denoting the cardinality of the set $ C(\bmV_{\text{set}})$.

\end{document}